\documentclass{article}

\usepackage{microtype}
\usepackage{graphicx}
\usepackage{subcaption}
\usepackage{booktabs} 
\usepackage{amsmath}
\usepackage{breqn}
\usepackage{url}
\usepackage{breakurl}
\usepackage[breaklinks]{hyperref}

\usepackage{adjustbox}
\usepackage{cleveref}

\usepackage[accepted]{icml2019}

\icmltitlerunning{INT8 Quantization of TransformerLT Model}

\begin{document}

\twocolumn[
\icmltitle{Efficient 8-Bit Quantization of Transformer Neural Machine Language Translation Model}



\icmlsetsymbol{equal}{*}

\begin{icmlauthorlist}
\icmlauthor{Aishwarya Bhandare}{int}
\icmlauthor{Vamsi Sripathi}{int}
\icmlauthor{Deepthi Karkada}{int}
\icmlauthor{Vivek Menon}{int1}
\icmlauthor{Sun Choi}{int}
\icmlauthor{Kushal Datta}{int}
\icmlauthor{Vikram Saletore}{int}
\end{icmlauthorlist}

\icmlaffiliation{int}{Artificial intelligence products group, Intel Corporation}
\icmlaffiliation{int1}{Work done while at Intel corporation}

\icmlcorrespondingauthor{Kushal Datta}{kushal.datta@intel.com}
\icmlcorrespondingauthor{Vikram Saletore}{vikram.saletore@intel.com}
\icmlcorrespondingauthor{Deepthi Karkada}{deepthi.karkada@intel.com}

\icmlkeywords{Machine Learning, ICML}

\vskip 0.3in
]



\printAffiliationsAndNotice{}  

\begin{abstract}
In this work, we quantize a trained Transformer machine language translation model to lower precision 8-bit integers. We leverage the high performance Intel\textsuperscript{\textregistered} Math Kernel Library martix multiplication kernels optimized with INT8/VNNI instructions in the latest Intel\textsuperscript{\textregistered} Xeon\textsuperscript{\textregistered} Cascade Lake processors to improve inference efficiency while maintaining less than 0.5 drop in BLEU score accuracy. To the best of our knowledge, this is the first attempt in the industry to quantize the Transformer model. We present novel quantization techniques directly in TensorFlow to opportunistically replace 32-bit floating point (FP32) computations with 8-bit integers (INT8) and transform the FP32 computational graph. We also present a parallel batching technique to maximize CPU utilization during inference. Our optimizations improved performance of both FP32 and INT8-quantized model resulting in a net improvement of 1.5X of the best quantized model over the best FP32 performance. Furthermore, we reveal opportunities and challenges of quantizing emerging deep learning model inference on Intel CPUs and establish best practices to do so.
\end{abstract}

\section{Introduction}
\label{sec:introduction}

The Transformer model using self-attention mechanism has recently achieved the state of the art accuracy in language translation \cite{Vaswani2017-rd}. Compared to its predecessors, this sequence transduction model circumvents recurrent or long-short memory (LSTM) neural cells and exploits multi-headed attention mechanism to capture global dependencies between input and output word sequences. Since its first use in machine translation, the multi-headed attention mechanism has shown tremendous promise in speech recognition \cite{DBLP:journals/corr/abs-1712-01769}, generative language modeling \cite{DBLP:journals/corr/abs-1709-08878}, machine reading comprehension \cite{DBLP:journals/corr/HuPQ17}, Language representation models \cite{bert-arxiv} and other natural language processing workloads.

The growing presence of machine language translation services and tools \cite{Microsoft2018}, \cite{Google2018}, \cite{AWS2018} and \cite{LingoTek2019} to name a few, clearly shows that machine translation inference is an important workload.
Quantization is a technique to improve the performance of inference workloads by using lower precision data types (8-bit, 4-bit or 2-bit integers) in place of 32-bit floating point. The latest Intel\textsuperscript{\textregistered} Xeon\textsuperscript{\textregistered} Cascade Lake processors include specialized vectorized neural network instructions (VNNI) to expedite quantized inference by fusing 64 8-bit multiply and add (FMA) operations into a single instruction \cite{VNNI2018}. This means that the vectorized FMAs can be completed in fewer clock cycles than previous generation Intel\textsuperscript{\textregistered} Xeon\textsuperscript{\textregistered} processors. As a result, 8-bit matrix multiplications (MatMuls) or quantized MatMuls execute faster on these platforms. This motivated us to explore the impact of VNNI on the performance of Transformer model inference. 
To the best of our knowledge, the Transformer model has not been quantized before. 
However, the impact of quantized MatMuls on the overall performance of Transformer inference was not known before this work as speedup between INT8 and FP32 MatMuls depend on the shape and size of the matrices involved.

Additionally, we want to minimize the drop in translation accuracy which can result due to the usage of reduced precision data types. In this work, our contributions include the following:
\begin{enumerate}
	\item Quantized a trained FP32 Transformer model to INT8 to achieve $<$ 0.5 drop in state-of-the-art (SOTA) BLEU score.
	\item Improve inference performance by:
	\begin{enumerate}
		\item Optimizing quantized MatMuls for tensor shapes and sizes in the Transformer model
		\item Reducing overhead due to quantization operations in the Transformer model compute graph
		\item Optimizing input pipeline by ordering sentences by token length
		\item Implementing parallel execution of batches with increased inference throughput
	\end{enumerate}
\end{enumerate}
The rest of the paper is organized as follows. In \autoref{sec:relatedworks}, we describe prior work on quantization techniques for deep learning models. In \autoref{sec:modeldesc}, we provide a brief description of the Transformer translation model. In \autoref{sec:quantization}, we describe how we first quantize the graph to maintain accuracy and then in \autoref{sec:performance} detail strategies to improve inference efficiency.
In \autoref{sec:results}, we present the overall improvement in inference performance.

\section{Related work}
\label{sec:relatedworks}

Various techniques including quantization i.e. using lower precision data types with smaller bit-widths have been proposed to compress deep neural network models. Wu et al used quantization to compressed and accelarate neural networks, but their work was restricted to only CNNs \cite{Wu2016CVPR}. Other related work on quantization proposed by \cite{DBLP:journals/corr/CourbariauxB16}, \cite{Ristretto}, \cite{Park_2018_ECCV} and \cite{Nvidia2017TensorRT} all have similar restrictions. \cite{DBLP:journals/corr/abs-1802-08635} proposed a ternerization scheme for quantization of weights to compress the neural network model, but didn't deal with quantization of activations. The authors in \cite{Naveen2017Ternary} quantized both weights and activations using a sub 8-bit inference pipeline where weights were constrained to +1, 0, -1 and activations were quantized using either 8 or 4-bits. Although, they reduced memory consumption, their technique failed to improve inference performance.\\
Binary, ternary and quaternany quantization techniques for recurrent methods including LSTM, GRU and ConvLSTM were proposed by the authors in \cite{AlomRNNQuant} . Although they observed promising performance results, they did not explore the applicability of their technique to the Transformer model with self-attention layers.
To the best of our knowledge, this is the first work where quantization has been applied to the Transformer language translation network. Transformer network does not contain convolutions or recurrent layers, only the attention mechanism making it unique. This work is also the first to demonstrate the speedup observed in inference performance by leveraging VNNI.

\section{Model Description}
\label{sec:modeldesc}

The architecture of the Transformer translation model with multi-headed attention is shown in figure 1 of \cite{Vaswani2017-rd}. It has an encoder-decoder structure. The encoder maps the input sequence of tokens in source language to a sequence of latent representations. The decoder generates the translated sequence of tokens in target language from these representations. The decoder is auto-regressive which means that previously generated tokens are used to decode the next token using a while loop. \\
The model uses the scaled dot-product attention and multi-headed attention mechanisms. The equations \autoref{eq:attention} and \autoref{eq:multihead} from \cite{Vaswani2017-rd} describe the fundamental computation in the model. 
It can be inferred from the equations that the primary operation in this model is a Matrix Multiplication (MatMul). It is also clear that the model contains the Softmax operation in between MatMuls. The Softmax operation in \autoref{eq:softmax} involves mapping the input to a probability distribution, and has a division operation. This would mean that computing a Softmax in a lower precision datatype would result in high accuracy loss as compared to computing it in full-precision FP32 datatype. In addition to Softmax, the graph has Layer Normalization\cite{lei2016layer} layer in between any two layers\cite{Vaswani2017-rd}. The Layer Normalization layer involves calculating the mean and variance of each layer, and normalizing the values of that layer. This involves operations like division, square and square root, which again are more accurate with a full-precision FP32 datatype over INT8 datatype.
Thus, the entire computation graph of this model doesn't support low precision INT8 datatype. Parts of the graph need to be selectively transformed to work in low-precision, while keeping the remainder of the graph in FP32.

\begin{equation}
Attention(Q,K,V) = Softmax(\frac{QK^T}{\sqrt{d_K}})V
\label{eq:attention}
\end{equation}

\begin{align}
	\begin{aligned}
		MultiHead(Q,K,V) &= concatenate(head_1,\\
		& \qquad head_2,\ldots,head_h)W^O
		\label{eq:multihead}
	\end{aligned}\\
	\begin{aligned}
		where, head_i = Attention(QW_1, KW_2, VW_3) \nonumber
	\end{aligned}
\end{align}


\begin{equation}
  \label{eq:softmax}
  Softmax(\phi_k) = \frac{\exp(\phi_k)}{\sum^{c}_j \exp(\phi_j)},
\end{equation}
Transformer model has two variations -- base and big. The differences are in the number of weight dimensions and number of layers. For English to German translation, the base and big model achieve BLEU scores of 27.3 and 28.4 respectively. In this study, we retrained the base model and start with a BLEU score of 27.68, which is close to the state-of-the-art.
\cite{TransformerOfficial}.

\section{Quantization with accuracy}
\label{sec:quantization}
Quantizing a model entails converting the FP32 tensors to a target lower precision integer data type as described in \autoref{eq:quantization} and then replace the FP32 operations with corresponding quantized operations.

\begin{gather}
	scale = \frac{target}{Max-Min} \\
	A_{quantized} = round((A_{float} - zero_{offset}) \cdot scale )
	\label{eq:quantization}
\end{gather}

For example, to convert an FP32 tensor following \autoref{eq:quantization} to an unsigned INT8 tensor the simplest way is to map minimum (Min) and maximum (Max) values in the tensor to 0 and 255 respectively. Values between 0 and 255 represent the numerical distribution between Min and Max. This way we can represent an arbitrary magnitude of ranges to a lower precision datatype. The overhead of quantization is $O(N)$ as it requires linear scans to calculate Min and Max, and then transform the entire tensor. $N$ here signifies the size of the tensor.

\subsection{Na\"ive Quantization}
\label{sec:naive}
\begin{figure}
    \centering
	\includegraphics[width=\linewidth]{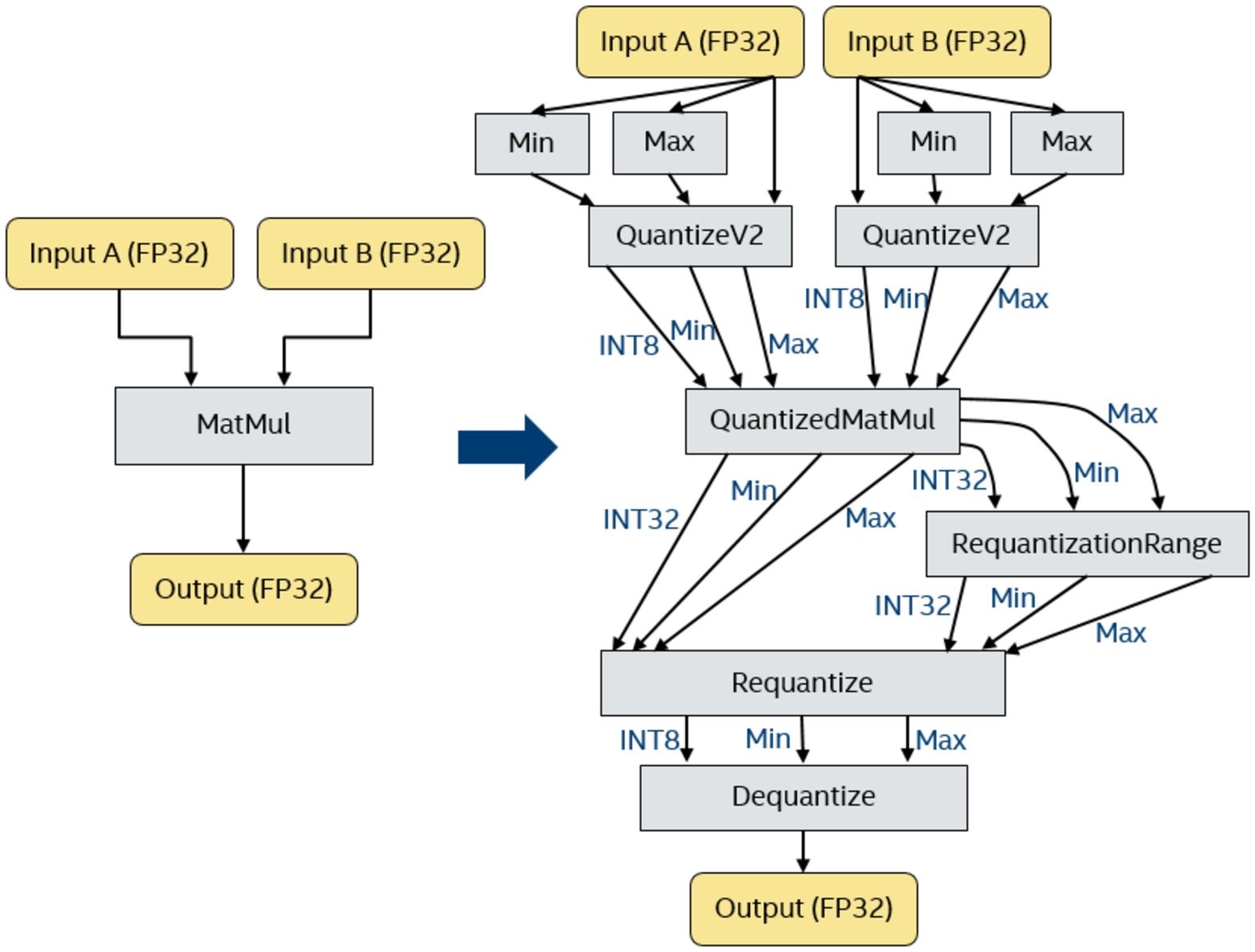}
	\caption{Schematic of Na\"ive Quantization for MatMuls in TensorFlow}
	\label{fig:matmul-naive-quant}
\end{figure}

The primary compute in neural networks is in the form of convolution and matrix multiplication operations. These operations benefit from quantization due to speed-up obtained from INT8/VNNI instructions. The Transformer LT model has MatMuls as seen in \autoref{sec:modeldesc}, and MatMuls take up the largest portion of the computation as shown in \autoref{fig:opsplits}(a). Thus, MatMul is the first operation that must be quantized. \\
The na\"ive way of quantization is mapping the entire FP32 dynamic range of a tensor to INT8. The modification in the computation graph in TensorFlow as shown in \autoref{fig:matmul-naive-quant} involves replacing a quantizable operation (MatMul) with the corresponding quantized version of that operation(QuantizedMatMul) in the model graph, and passing quantized input tensors to that operation. The quantized tensors are obtained by passing the FP32 tensor through a QuantizeV2 operation to convert it to INT8. Operations to calculate Min and Max values of the tensor are also inserted before the QuantizeV2 node. The difference between quantized kernel of an operation and its FP32 version is that it executes ISA optimized INT8 instructions (including VNNI). For example, the QuantizedMatMul operation accepts the A matrix (signed INT8), the B matrix (unsigned INT8) and their corresponding Min/Max values. The result of the multiplication is accumulated as signed INT32 value. The RequantizationRange operation calculates the output Min and Max values from the INT32 result. This is fed into a Requantize operation which maps the INT32 result into an INT8 representation with a new Min and Max value. Since operations downstream in the computational graph may not accept INT8 datatype values, a Dequantize operation is required to convert the INT32 value back into an FP32 value as shown in Equation \ref{eq:dequantization}. Dequantization is also $O(N)$ in complexity.

\begin{dmath}
\label{eq:dequantization}
A_{dequantized}=(Max-Min) \cdot (A_{quantized}-zero_{offset})
\end{dmath}

This transform was done to all the MatMul operations in the graph. On running inference on the graph transformed using this method, it failed to emit a stop token at all, and consequently failed to achieve the less than 0.5 drop in the FP32 BLEU score accuracy. This ended up deeming the na\"ive quantization approach inappropriate for this model. A closer investigation by visualizing the histograms of input tensor values show that most of them have a long-tailed numerical distribution.

\begin{figure*}[t]
	\centering
	\includegraphics[width=\textwidth]{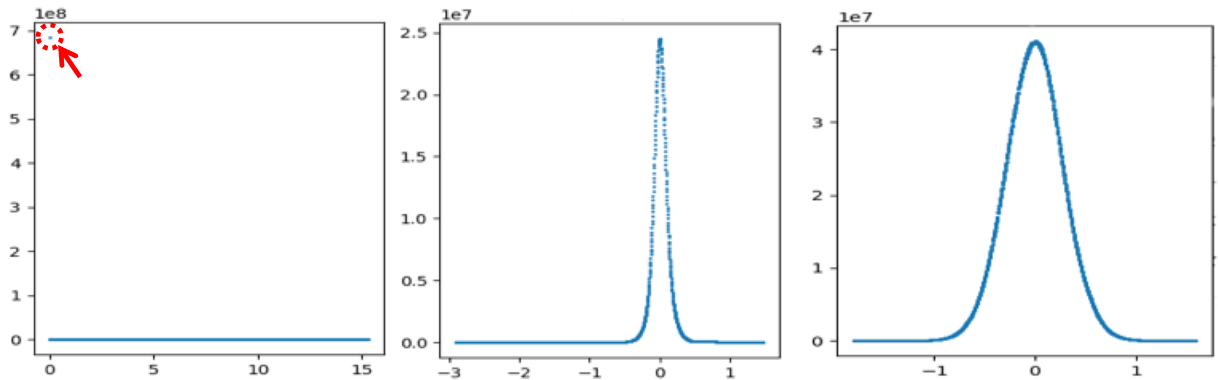}
	\caption{Tensors with sparse, narrow and Gaussian histograms in Transformer model}
	\label{fig:histograms}
\end{figure*}

If the entire range of the FP32 tensor was to be preserved, there would be a loss of precision due to multiple values being mapped to the same bin. 
 Hence, quantization using absolute Min and Max can result in significant loss in accuracy. As a result na\"ive quantization is not a viable approach. In the next subsection, we explore other approximation methods based on divergence between probability distributions.

\subsection{KL-Divergence for optimal saturation thresholds }
\label{sec:kldivergence}

One solution to preserve precision when quantizing tensors is to reduce the range of the representation. This relies on the assumption that maintaining small differences between tensor values that are close together is more important than representing the absolute extreme values or the outliers. Ideally, the numerical distribution of values in the mapped INT8 tensor representations should be as close as possible to the distribution of values for FP32 tensors. One way to measure this ``closeness`` is to use the Kullback-Leibler Divergence (KL-Divergence) \cite{10.2307/2236703} metric between the histograms of FP32 and INT8 tensors. By iteratively choosing different Min and Max threshold values and mapping them to their respective extrema in the INT8 representation, we are able to find optimal Min and Max values that minimize the KL divergence between the INT8 and FP32 tensors. We refer to this process as “calibration” within the quantization workflow. This idea was first introduced in \cite{Nvidia2017TensorRT}.

We chose 600 random length samples out of 3003 sentences in the validation dataset as calibration data for quantization. The MatMul input tensors in Transformer LT graph come from three types of distributions, as classified from inspection of histogram values from inferring on the calibration dataset shown in \autoref{fig:histograms}. 
The sparse tensors, when quantized, result in unacceptable accuracy degradation, and are kept unquantized i.e. with FP32 data type. 
For the other two types of tensors, they can be thresholded to get a reasonable accuracy degradation. The inputs to the MatMul in this model are both signed FP32, as opposed to the expected case of one signed weight and an unsigned activation. Thus, there is a need to quantize one of the tensors to unsigned INT8, which is described in detail in \autoref{sec:optimizeMatMuls}. The way to generate the thresholds for this conversion of a signed FP32 tensor to unsigned INT8 tensor affects the overall accuracy of the model. We determined the positive and negative thresholds using three separate ways, referred to as quantization modes:
\begin{enumerate}
	\item  \text{Symmetric} calculates the KL-divergence on the entire distribution. Here, \\
	$$Threshold_{Min}=-Threshold_{Max}$$
	\item  \textit{Independent} separates the distribution about value zero and calculates $Threshold_{Min}$ and $Threshold_{Max}$  independently
	\item  \textit{Conjugate} separates the distribution about zero and calculates thresholds independently, but reports \\
	$$\left\{\begin{matrix}
    Threshold_{Max}=\max(|Max|,|Min|)\\ 
    Threshold_{Min}=-Threshold_{Max}
    \end{matrix}\right.$$
\end{enumerate}

\begin{table}[t]
	\centering
	\caption{Effects of calibration modes on the accuracy}
	\label{Tab:modes}
	\begin{adjustbox}{max width=\columnwidth}
	\begin{tabular}{|l|c|c|}
		\hline
		Mode & BLEU score & Drop in Accuracy \\
		\hline \hline
		Na\"ive quantization & NA & NA \\
		\hline
		Symmetric & 27.30 & 0.38 \\
		\hline
		Independent & 27.33 & 0.35 \\
		\hline
		Conjugate & 27.26 & 0.421 \\
		\hline
	\end{tabular}
	\end{adjustbox}
\end{table}

\autoref{Tab:modes} shows the effect of different quantization modes on the final acuracy (BLEU score). One test of the need for early thresholding using KL-divergence is to perform na\"ive quantization on all the MatMuls that do not have a sparse input. The graph generated by this method could not emit a STOP token during inference, giving out garbage translations. The BLEU score is unavailable, marked as NA in the table. This proves the need to use early thresholding.
It can also be observed that independently calculating the thresholds for positive and negative halves of the histogram results in the least drop in accuracy. In this case, the min and max thresholds might not be symmetric about zero, causing the quantized tensors to have a non-zero value for the offset. This results in the QuantizedMatMul kernel being slightly slower than the case where the offsets are zero.
Thus, we use the ‘symmetric’ mode for threshold calculation at a small cost of 0.03 drop in accuracy. Note that we ended up not quantizing the tensors with Sparse histogram (appearing in 12 out of 97 MatMuls). Tensors with narrow and Gaussian distributions are quantized. 

\section{Improving Performance}
\label{sec:performance}

Our objective here is to not leave any performance on the table; exploit every opportunity to improve efficiency including (but not limited to) leverage MatMul kernels optimized with VNNI instructions, reduce operation times in the model, fuse operations, optimize input pipeline, improve resource utilizing with parallelization and other strategies. In the process, we reduced total number of operations in the graph by removing redundant operations and reordering operations maintaining correctness. We introduced new optimizations in the Intel\textsuperscript{\textregistered} MKL kernel implementations, found a new way of ordering input sentences and parallelized execution of batches. These techniques are described in the following subsections.

\subsection{Optimizing Quantized MatMuls}
\label{sec:optimizeMatMuls}

\begin{figure}
\centering
\includegraphics[scale=0.5]{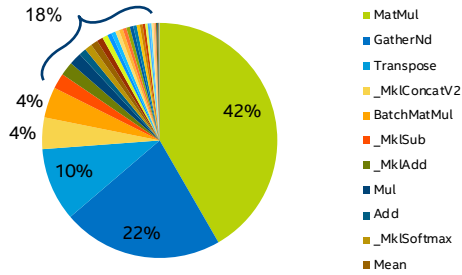}
\caption{Operation times of FP32 Transformer Model}
\label{fig:fp32optimes}
\end{figure}

\begin{figure}
	\centering
	\includegraphics[scale=0.25]{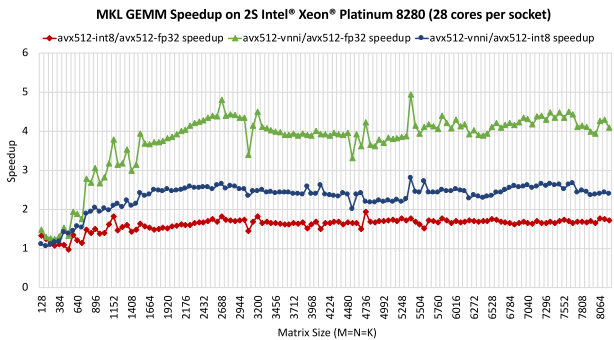}
	\caption{MKL GEMM Speedup with INT8/VNNI compared to AVX512 with micro-benchmarks}
	\label{fig:gemm1}
\end{figure}

Earlier generation (before 2019) Intel CPUs designed with 512-bit Advanced Vector Extensions instructions (AVX512) helped vectorize both 16 FP32 and 64 INT8 multiply-and-add operations. 2nd Generation Xeon\textsuperscript{\textregistered} Scalable codenamed Cascade Lake CPUs designed with INT8/VNNI instructions further optimized them. The figure \autoref{fig:gemm1} shows INT8 MatMuls using VNNI provides a speed-up of 3.7X over FP32 MatMuls using AVX512. And, the speed-up provided by VNNI over AVX512 for INT8 MatMul is 2.3X. In this context, as seen in the operation splits in \autoref{fig:fp32optimes}, MatMuls take 42\% of time in the FP32 model. This data strongly motivates us to use VNNI instructions to expedite MatMuls.

The default TensorFlow version 1.12 use open-source GEMMLOWP library for the integer matrix-matrix multiplication \cite{GemmLowP}. When we started this work, GEMMLOWP kernel did not use INT8/VNNI instructions. Additionally, GEMMLOWP required format conversions of the input matrices prohibitively reducing its efficiency on our platform. Quantized MatMul kernel in Intel\textsuperscript{\textregistered} MKL BLAS, on the other hand, is IA optimized and exploit INT8/VNNI goodness. Hence, as our first order of business we integrated MKL integer GEMM (General Matrix Multiply) kernel to TensorFlow INT8 quantized MatMul which accepts one signed and one unsigned INT8 matrix and accumulates results into a signed INT32 tensor.

\begin{figure}
\includegraphics[scale=0.25]{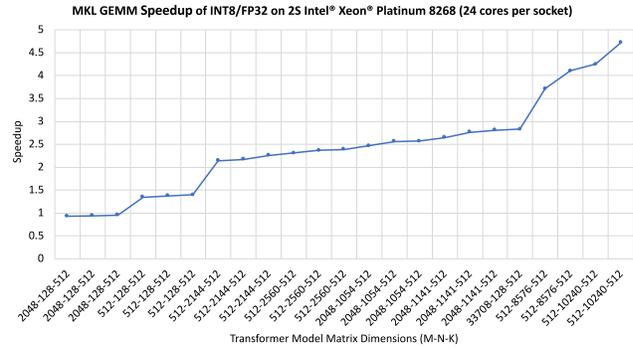}
\caption{Comparison of speedups of MKL GEMM INT8 vs FP32 for different matrix shapes}
\label{fig:gemm2}
\end{figure}

However, our first implementation resulted in lower than expected performance boost. Further analysis showed that the MatMul shapes used in the Transformer model were running with lower efficiency due to non-zero offsets. To elaborate, Equations \ref{eq:matmul1} and \ref{eq:matmul2} below shows how a single call to GEMM\_S8U8S32: A (INT8), B(UINT8), C(INT32) implicitly executes 6 MatMuls with non-zero offsets. In the equations, $\alpha$ and $\beta$ are scalar values, $oa$, $ob$ and $oc$ are offsets for A, B and C respectively. While the first operation i.e. $Op(A)Op(B)$ is optimized in the MKL kernel, the other multiplications were not. Fixing this gap resulted in an average 2.4X speedup across all the MatMuls in the model as shown in \autoref{fig:gemm2}. We consider this one of our key achievements.

\begin{dmath}
\label{eq:matmul1}
C = \alpha(Op(A) + oa)(Op(B) + ob) + \beta C + oc
\end{dmath}

\begin{dmath}
\label{eq:matmul2}
C = \alpha(Op(A)Op(B) + Op(A)ob + Op(B)oa + oa*ob) + \beta C + oc
\end{dmath}




\subsection{Optimizing GatherNd}
\label{sec:optimizeGatherNd}

The GatherNd operation on N-dimensional tensors uses the input indices to perform a Gather on the input tensor to form the output tensor. In total, there are 40 such GatherNd operations in the Transformer model which occur in the decoder while loop. These operations involve copy on large tensors, making it one of the most expensive operations due to beam search. There is no obvious compute benefit for quantizing GatherND. Looking into the GatherNd implementation in TensorFlow, we found that memory copy time dominates the operation time. Hence, we reduced amount of data copied by changing the data type of the tensors to INT8 via quantizing GatherNd.

\begin{figure}
	\includegraphics[width=0.5\textwidth]{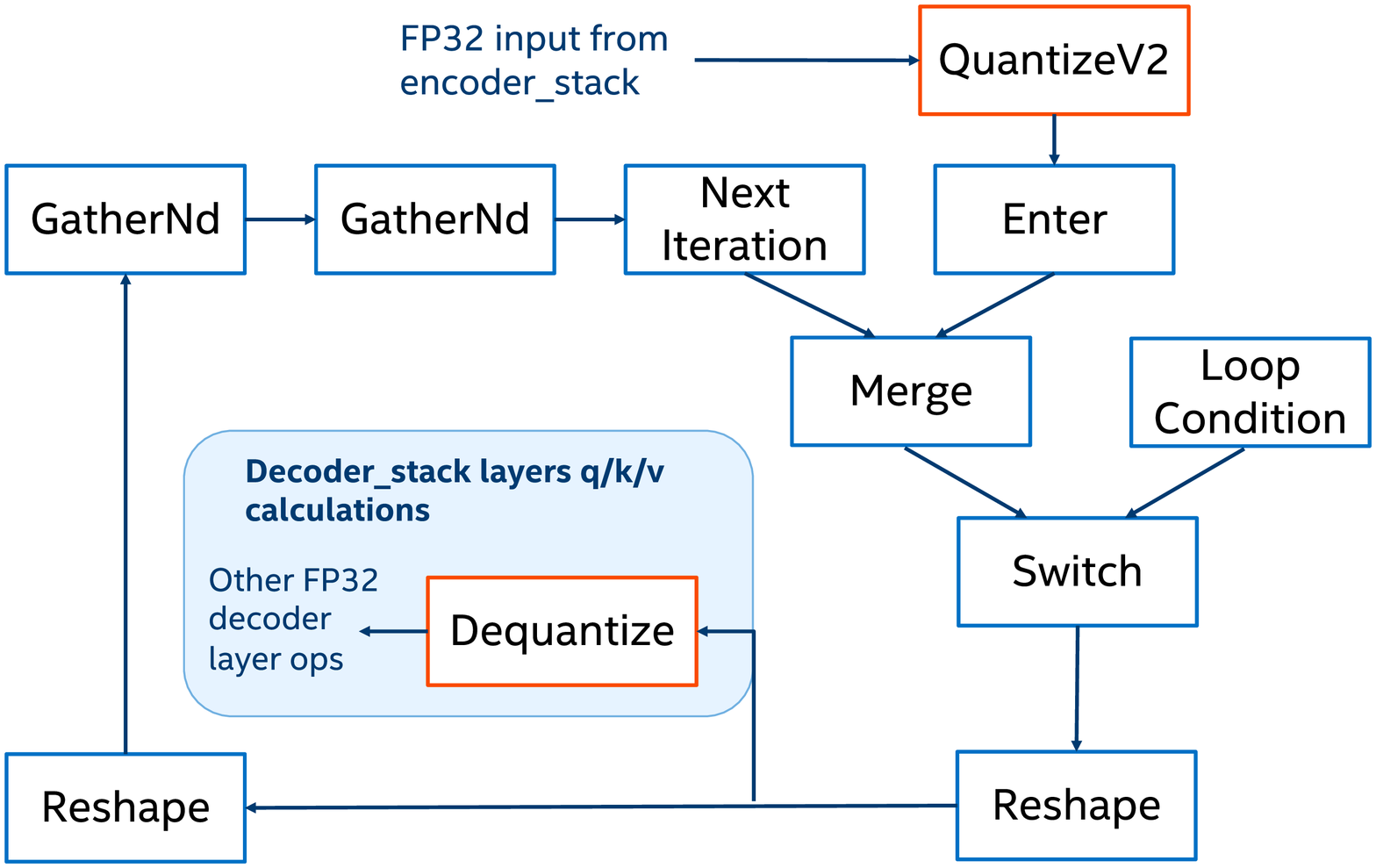}
	\caption{Operations in the while loop of the Transformer model}
	\label{fig:gathernd}
\end{figure}

In the Transformer computational graph, the decoder while loop contains GatherNd operations as shown in \autoref{fig:gathernd}. The na\"ive way of quantizing GatherNd involves adding a Quantize and Dequantize node before and after the GatherNd node. This, however adds an overhead, which reduces the overall speedup. We however managed to minimize the extra cost of this Dequantize by repositioning the existing the Quantizes and Dequantizes in the graph due to QuantizedMatMul. With these changes we reduced the copy size by 3.8X for the validation dataset. The execution time of the GatherNd operation alone was reduced by 5X.

\subsection{Sorting Input Sentences}
\label{sec:sorting_inputs}
In machine translation, the inputs to the network have varying sequence lengths. When input
sentences are batched together, all the sentences except the longest sentence in the batch are padded to the sequence length of the longest sentence in each batch. This adds an overhead per batch in terms of wasted computations for the pad tokens. To work around this issue, it is important to sort the input validation dataset to keep the padding added to each batch to a minimum. There are different sorting methodologies such as sorting based on the number of words in each input sentence or token sorting based on the number of tokens in each input sentences. We have found that inference performance with sorting based on the number of tokens gives us an improvement of 28\% over inference performance with sorting based on the word count of the input sentence.

\begin{figure}
    \centering
	\includegraphics[scale=0.5]{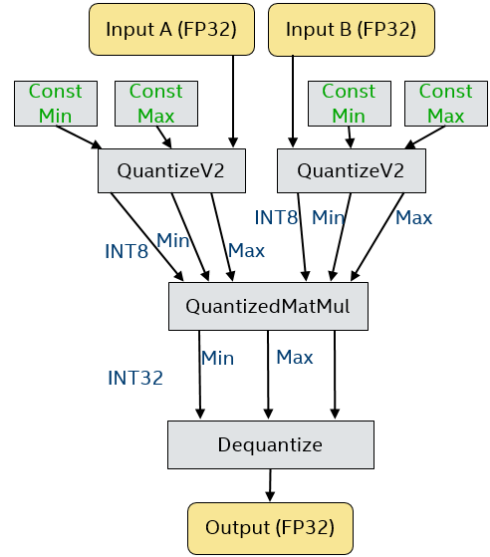}
	\caption{Schematic of Optimized Quantization for MatMuls in TensorFlow}
	\label{fig:afteroptgraph1}
\end{figure}

\subsection{Eliminating Operations from Graph}
\label{sec:eliminateOperations}

As discussed in \autoref{sec:kldivergence}, finding thresholds using KL-divergence method eliminated the need for computing absolute Min and Max of tensors in the graph. These threshold values are inserted as Const operations in the graph. This further removed some of the Reshape operations. We also eliminated Requantize and RequantizationRange operations for tensors which were feeding in to unquantized operations. We used a Dequantize operation to convert INT32 to FP32 directly as shown in \autoref{fig:afteroptgraph1}. These removals contributed to reducing the total number of operations in the quantized compute graph. Additional quantize/dequantize operations were also removed in the GatherNd quantization.

\begin{figure}
	\includegraphics[width=0.5\textwidth]{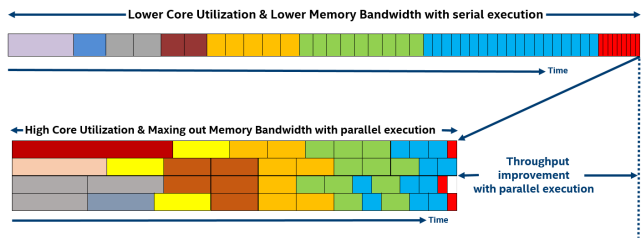}
	\caption{Comparison of the serial and parallel execution techniques}
	\label{fig:parallelbatching}
\end{figure}

\begin{figure*}
	\centering
	\begin{subfigure}[!h]{0.9\textwidth}
		\includegraphics[width=\textwidth]{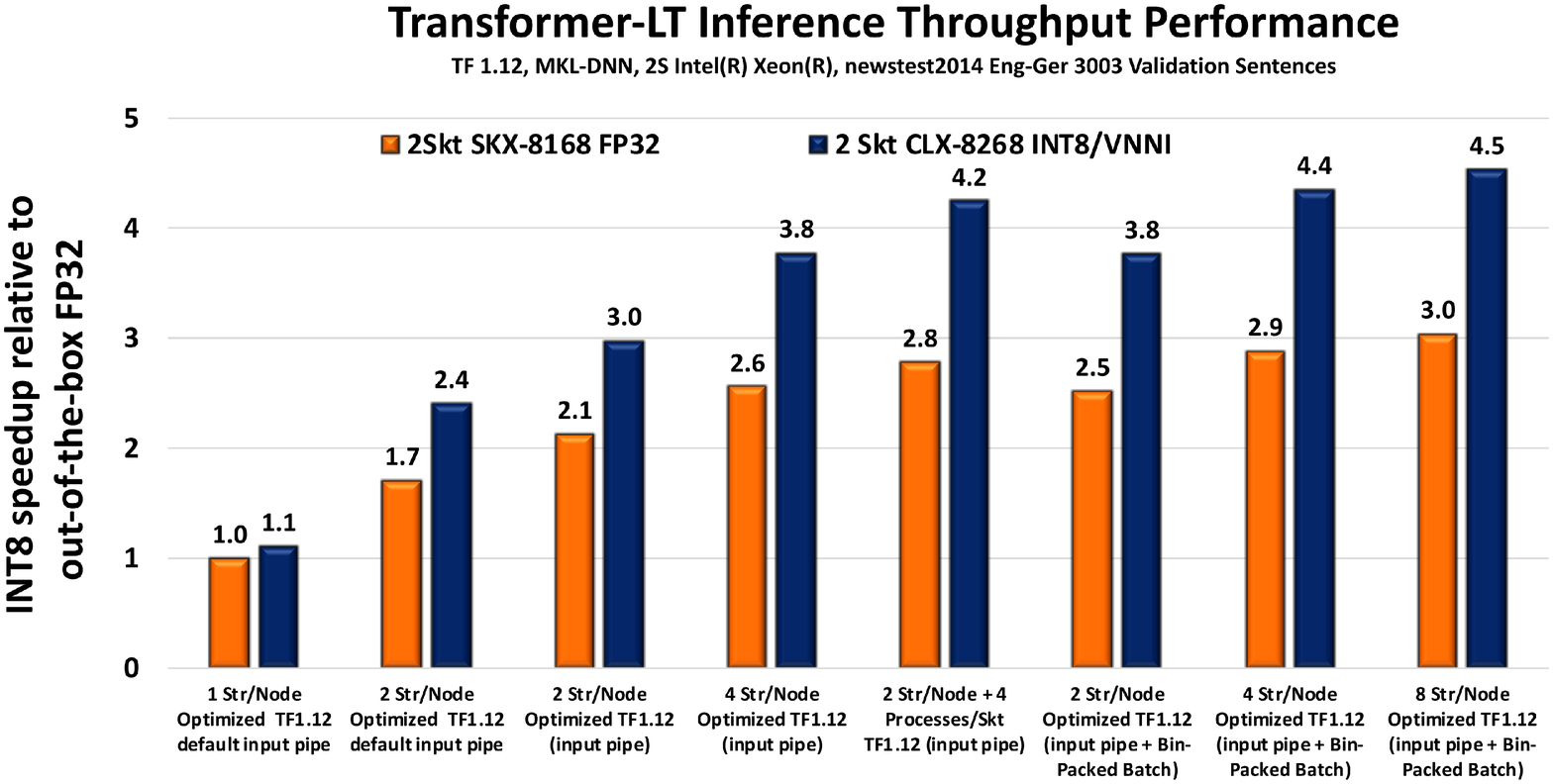}
		\caption{}
		\label{fig:finalresults_1}
	\end{subfigure}
	
	\begin{subfigure}[!h]{0.9\textwidth}
		\includegraphics[width=\textwidth]{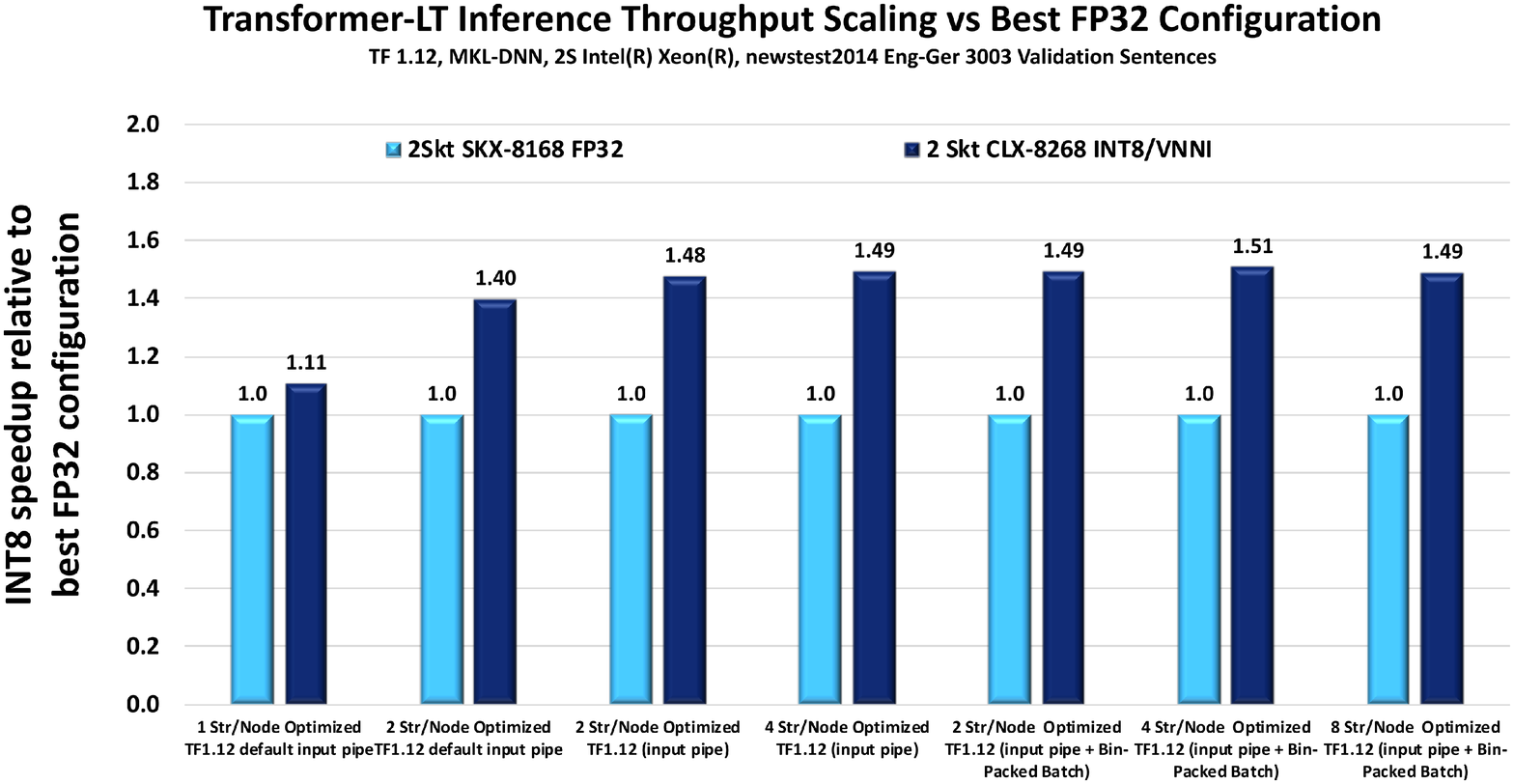}
		\caption{}
		\label{fig:finalresults_2}
	\end{subfigure}
	\caption{Throughput Performance of INT8/VNNI on 2S Xeon\textsuperscript{\textregistered} 8268 vs FP32 2S Xeon\textsuperscript{\textregistered} 8168}
	\label{fig:finalthroughput}
\end{figure*}

\begin{figure*}
    \centering
	\includegraphics[width=0.7\textwidth]{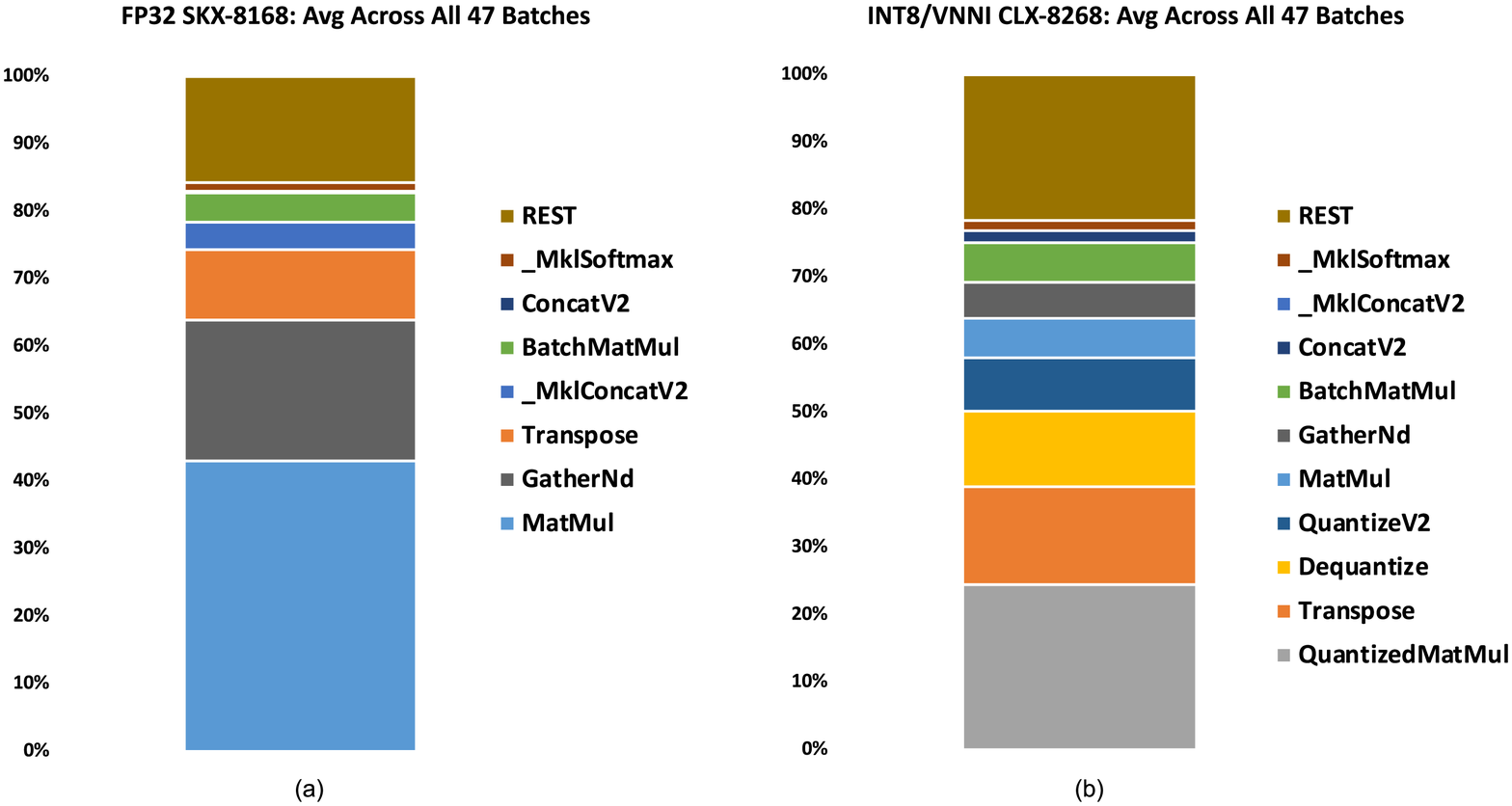}
	\caption{Distribution of percentage operation times in FP32 vs INT8 graph; INT8/VNNI avg time per batch is lower than FP32}
	\label{fig:opsplits}
\end{figure*}

\subsection{Parallel Batching}
\label{sec:binpacking}

The execution time of inference varies depending on the length of the sentences in the batch. This occurs due to larger number of operations (such as Matmuls) with increased sentence length and decode steps. Our measurements repeatedly showing that CPU utilization significantly drops depending on sentence size led us to investigate how we could further exploit this to expedite performance. Since batches of longer sentences more efficient use CPU cores, serially executing batches in-order seemed inefficient. We could pack one or more batches of longer sentences with batches of shorter sentences and effectively increase CPU utilization.

To achieve this, we create a parent TensorFlow session which in turn creates a FIFO batch queue. This parent process creates children processes which are affinitized to specific subset of CPU cores and local memory (NUMA) domain. The children processes pick batches of input sentences from the batch queue and perform inference. The input sentences are ordered by decreasing token count before being added to the batch queue. Note that the child processes dequeue batches asynchronously from the batch queue. This means batches of long and short sentences are processed in parallel utilizing the cores more efficiently. As a result of this optimization, we observe a 1.4X improvement in throughput as shown in \autoref{fig:parallelbatching}. We use multiple inference streams per node as described in \cite{MultiWorker}.

\section{Throughput Performance Results}
\label{sec:results}

\textbf{Experimental Setup:} FP32 and INT8 performance are evaluated on 2S Intel\textsuperscript{\textregistered} Xeon\textsuperscript{\textregistered} Platinum 8168 (24 cores per socket) processors and Intel\textsuperscript{\textregistered} Xeon\textsuperscript{\textregistered} Platinum 8268 (24 cores per socket) processors, respectively. Both are tested with TensorFlow 1.12.0 built with VNNI, Intel\textsuperscript{\textregistered} MKL mklml$\_$lnx$\_$2019.0.3.20190119, Python 2.7 on CentOS 7.5.
The inference performance of Transformer model for both INT8 and FP32 graphs are evaluated with newstest2014 dataset. A mini batch size of 64 is used in all experiments. In \autoref{fig:opsplits}, percentage of operation times are shown in different colors. MatMul is the major operation that accounts for 43\% of the FP32 execution time. In case of INT8/VNNI quantized graph, some of these MatMuls are replaced with QuantizedMatMuls reducing the percentage of time spent in matrix multiplications. However, the quantization of MatMuls resulted in overheads such as Dequantize and QuantizeV2 in the INT8 graph. GatherND, another operation that took up a significant portion of the FP32 computation, also significantly reduced its INT8 percentage through the optimization described in \autoref{sec:optimizeGatherNd}.

\autoref{fig:finalthroughput} compares the overall inference throughput of the Transformer model with INT8/VNNI optimizations on 2S Xeon\textsuperscript{\textregistered} 8268 with optimized FP32 on 2S Xeon\textsuperscript{\textregistered} 8168 platforms. The first two bars in \autoref{fig:finalthroughput} are throughput results obtained by using default word-sorted input data measured with 1 and 2 streams per node respectively. The next two sets use token-sorted input data on 2 and 4 streams per node respectively. The last three sets use both token-sorted and parallel batching with 2, 4, and 8 streams per node respectively. \autoref{fig:finalresults_1} shows that we were able to achieve up to 4.5X throughput performance scaling with INT8/VNNI quantization relative to the out-of-the-box FP32 performance with all of our optimizations. However, our input pipeline and system level optimizations effectively improved FP32 performance by 3X.
\autoref{fig:finalresults_2} shows throughput performance scaling using INT8/VNNI relative to best FP32 SKX-8168 with the best system configuration. The highest INT8 throughput is achieved with 2 inference streams/socket with token sorting and parallel batching resulting in a scaling of 1.51X.

\section{Conclusion}
\label{sec:conclusion}
In this work we have quantized the Transformer machine language translation model in TensorFlow and maintained less than 0.5 drop  in  BLEU score accuracy. The key learning is that models using non-linear layers like Softmax and Layer Normalization appearing between layers like MatMul make the quantization process effort-intensive. For our readers who wish to attempt the same, our recommendation is to analyze distributions of FP32 values in the tensors and selectively quantize to achieve both high accuracy and speedup. Although this proof-point has been developed on Intel CPUs, we profess that the method can be generally applicable to all hardware platforms and all models that use self-attention or multi-head attention layers. It is essential to ensure that accuracy target is met while trying to improve the performance. We optimized the compute graph by reducing number of operations, improved kernels of key operations such as MatMuls and GatherNd, optimized order of sentences in the input pipeline and finally used parallel batching to achieve the highest throughput gains of 1.5X.

\section*{Acknowledgements}
We acknowledge the contributions from the  Intel\textsuperscript{\textregistered} TensorFlow Direct Optimization team and the Intel\textsuperscript{\textregistered} Math Kernel Library team.


\bibliography{example_paper}
\bibliographystyle{icml2019}

\section*{Notices and Disclaimers}
Performance results are based on testing as of dates shown in configuration and may not reflect all publicly available security updates. No product can be absolutely secure. See configuration disclosure for details. Optimization Notice: Intel's compilers may or may not optimize to the same degree for non-Intel microprocessors for optimizations that are not unique to Intel microprocessors. These optimizations include SSE2, SSE3, and SSSE3 instruction sets and other optimizations. Intel does not guarantee the availability, functionality, or effectiveness of any optimization on microprocessors not manufactured by Intel. Microprocessor-dependent optimizations in this product are intended for use with Intel microprocessors. Certain optimizations not specific to Intel micro-architecture are reserved for Intel microprocessors. Please refer to the applicable product User and Reference Guides for more information regarding the specific instruction sets covered by this notice. Software and workloads used in performance tests may have been optimized for performance only on Intel microprocessors.  Performance tests, such as SYSmark and MobileMark, are measured using specific computer systems, components, software, operations and functions.  Any change to any of those factors may cause the results to vary.  You should consult other information and performance tests to assist you in fully evaluating your contemplated purchases, including the performance of that product when combined with other products. For more complete information visit:  http://www.intel.com/performance.

\end{document}